\pdfoutput=1
\documentclass[final]{cvpr}

\usepackage{times}
\usepackage{epsfig}
\usepackage{graphicx}
\usepackage{amsmath}
\usepackage{amssymb}
\usepackage{multirow}
\usepackage{verbatim}
\usepackage{mathtools}
\usepackage{breqn}
\usepackage{float}
\usepackage{subcaption}
\usepackage{resizegather}
\usepackage{rotating}
\usepackage{enumitem}
\usepackage[normalem]{ulem}
\usepackage[numbers,sort]{natbib}

\usepackage[pagebackref=true,breaklinks=true,letterpaper=true,colorlinks,bookmarks=false]{hyperref}

\begin{document}


\title{Learning Monocular Depth Estimation via \linebreak Selective Distillation of Stereo Knowledge}

\author{Kyeongseob Song\\
Hyundai Motor Company, Korea\\
{\tt\small ks.song@hyundai.com}
\and
Kuk-Jin Yoon\\
Visual Intelligence Lab., KAIST, Korea\\
{\tt\small kjyoon@kaist.ac.kr}
}

\maketitle

\begin{abstract}
Monocular depth estimation has been extensively explored based on deep learning, yet its accuracy and generalization ability still lag far behind the stereo-based methods. To tackle this, a few recent studies have proposed to supervise the monocular depth estimation network by distilling disparity maps as proxy ground-truths. 
However, these studies naively distill the stereo knowledge without considering the comparative advantages of stereo-based and monocular depth estimation methods.  
In this paper, we propose to selectively distill the disparity maps for more reliable proxy supervision. Specifically, we first design a decoder (MaskDecoder) that learns two binary masks which are trained to choose optimally between the proxy disparity maps and the estimated depth maps for each pixel. 
The learned masks are then fed to another decoder (DepthDecoder) to enforce the estimated depths to learn from only the masked area in the proxy disparity maps. Additionally, a Teacher-Student module is designed to transfer the geometric knowledge of the StereoNet to the MonoNet.
Extensive experiments validate our methods achieve state-of-the-art performance for self- and proxy-supervised monocular depth estimation on the KITTI dataset, even surpassing some of the semi-supervised methods. 

\end{abstract}


\begin{figure}[!t]
\begin{center}
	\includegraphics[width=1.0\linewidth]{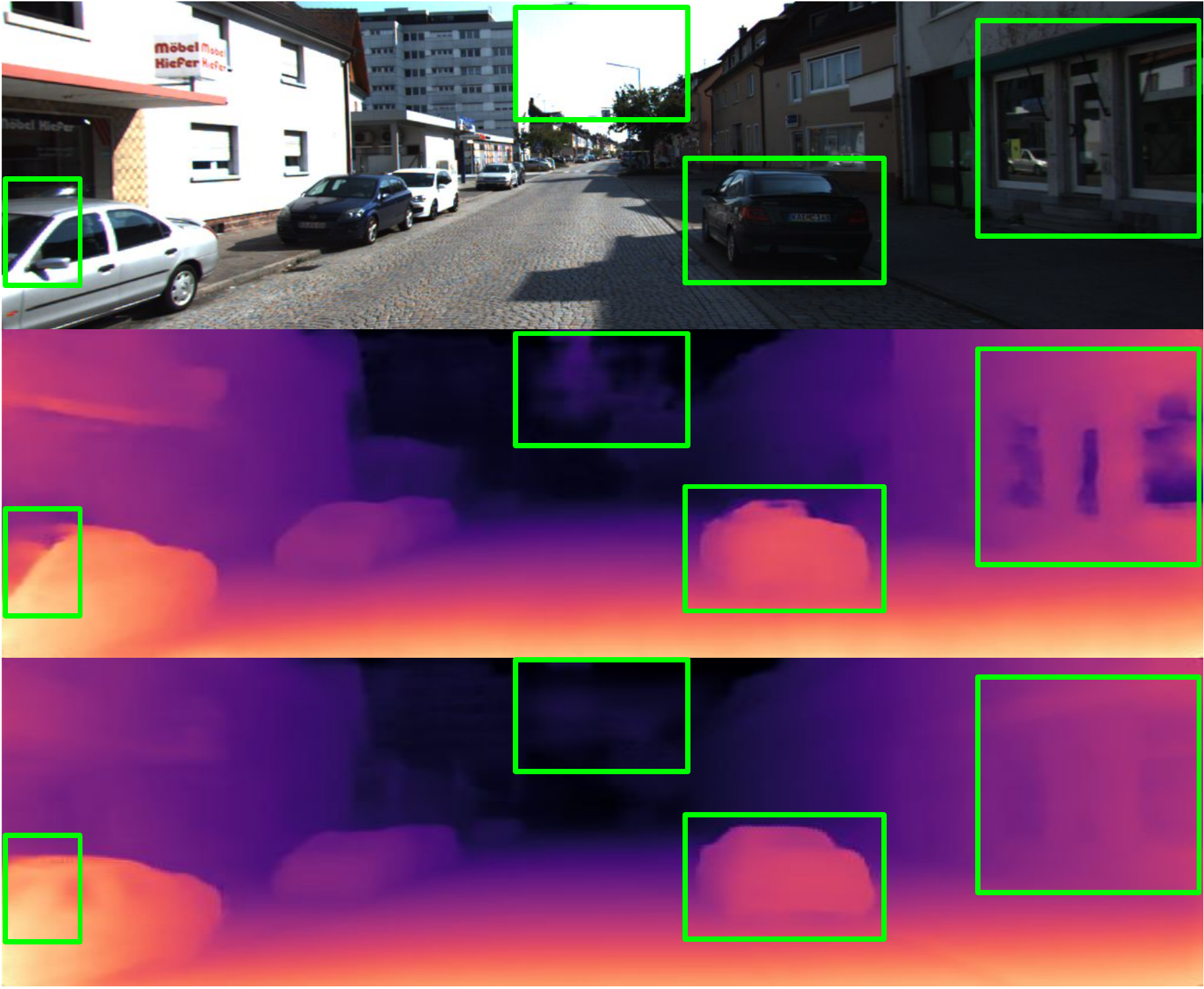}
\end{center}
    \vspace{-15pt} \captionsetup{font=small}
	\caption{Comparison of stereo and monocular depth estimation on one (\textit{top}) of the images from the Eigen test split of the KITTI dataset~\cite{geiger2013vision}. Corresponding disparity maps are predicted by \textit{a stereo network (middle)} from Guo \etal~\cite{guo2018learning} and \textit{our monocular depth estimation network (bottom)}, respectively.} 
	\label{fig:test_40}
    \vspace{-10pt}
\end{figure}

\vspace{-8pt}
\section{Introduction}
\label{sec:introduction}
Depth estimation is one of the pivotal 
computer vision tasks widely used in many applications such as 3D reconstruction, autonomous driving, augmented reality, etc. Among several methods of obtaining depth information from images, monocular depth estimation aims to infer the depth from a single image. Thanks to the wide applications, low cost and small size of monocular cameras, monocular depth estimation has been extensively studied along with the recent development of deep learning and convolutional neural networks(CNNs). However, one of the biggest challenges of supervised monocular depth estimation~\cite{alhashim2018high, lee2019monocular, gan2018monocular, fang2020towards, lee2019big, fu2018deep, zhao2019geometry} is that it requires a huge amount of expensive ground-truths. Utilizing synthetic data might be a possible solution for this issue, but training a monocular depth estimation network with synthetic datasets and evaluating it on real datasets is 
less applicable due to the large domain gap between the synthetic and real data. Alternatively, self-supervised monocular depth estimation methods~\cite{godard2017unsupervised, poggi2020uncertainty, godard2019digging, poggi2018learning, johnston2020self,  zhao2019geometry, gordon2019depth, chen2019towards, garg2016unsupervised} have been 
actively explored to reduce the photometric error between the reference image and the depth-projected image. However, as the depth projection is based on the view synthesis of the input images, it is sensitive to the input domain changes. Moreover, the accuracy is often less comparable to that of the supervised methods due to the inherent ambiguity of self-supervised loss functions~\cite{watson2019self}.

To alleviate the aforementioned issues, a few recent works~\cite{guo2018learning, tosi2019learning, watson2019self} have been proposed to use the disparity maps predicted by the stereo networks or traditional stereo matching methods such as Semi Global Matching (SGM)~\cite{hirschmuller2005accurate} as proxy ground-truths. 
They assume that the stereo-based methods generally provide better accuracy and have better generalization ability to other domains, to train monocular depth estimation networks. 
However, it is possible that monocular depth estimation could perform better than stereo-based methods in some areas.
For instance, as shown in  Fig.~\ref{fig:test_40}, 
stereo vision does not properly provide depth information in some areas, \eg, the occluded, textureless and reflective areas; while
monocular depth estimation methods suffer less from such problems in these areas as it requires only a single view. 
Conversely, monocular depth estimation is ill-posed since the size of objects is ambiguous, which can be improved by geometric cues from disparity maps. The problem is that earlier studies do not take account of these crucial considerations.

Inspired by these studies, 
this paper proposes a novel 
approach by transferring stereo knowledge to the monocular depth estimation network, in consideration of the comparative advantage of stereo and monocular depth estimation methods. 
A stereo network that is pre-trained on a large amount of synthetic data is first adopted as a proxy supervisory network for learning monocular depth estimation. 
The pre-trained stereo network predicts the proxy disparity maps given a pair of stereo images. At this step, unlike other works~\cite{guo2018learning, tosi2019learning, watson2019self} that directly leverage the disparity maps as proxy ground-truths, we design a novel decoder (MaskDecoder) which learns binary masks to selectively distill the proxy disparity maps. 
The learned masks indicate whether the proxy disparity maps provide superior pixel-wise estimation to the currently-estimated monocular depth maps or not. In other words, they imply the estimation can be improved if the masked pixels are guided by the proxy disparity map. To be more specific, every pixel in a learned binary mask selects either the proxy disparity(1) or the estimated depth(0) to define a dense virtual disparity map. The virtual disparity map is then optimized by the image reconstruction loss and edge-aware smoothness loss functions, which is equivalent to optimizing the selection of the binary masks. We elaborate the details in Sec.~\ref{sec:proposed_methods} and verify the selective distillation leads to superior accuracy to the direct distillation method.
Moreover, as the MonoEncoder extracts features from only a single image, we further improve the accuracy of the monocular depth estimation by employing a novel Teacher-Student (T-S) module \cite{wang2020knowledge} between the both encoders of stereo and monocular networks. The T-S module aims to transfer the geometric knowledge from the StereoEncoder (Teacher) to the MonoEncoder (Student).  
We analyze the effectiveness of each proposed method and evaluate our framework on the KITTI dataset~\cite{geiger2013vision}. 
The experimental results show that the proposed methods achieve state-of-the-art performance compared to other self-, proxy- and semi-supervised monocular depth estimation networks. 

\section{Related work}

In this section, we review the relevant stereo and monocular depth estimation literature and introduce several works that distill proxy supervision into the depth estimation.

\vspace{5pt}

\noindent\textbf{\noindent{Stereo depth estimation}} \hspace{1mm}
Stereo depth estimation aims to recover depth information by computing disparity from the correspondence across two images. According to~\cite{scharstein2002taxonomy}, the traditional stereo depth estimation follows a typical pipeline consisting of four steps: matching cost computation, cost aggregation, optimization and disparity refinement.
In this pipeline, CNNs~\cite{zagoruyko2015learning, chen2015deep, zbontar2016stereo, luo2016efficient} have been usually utilized to compute matching costs between two sampled patches, replacing the conventional steps. However, since these methods rely on patch-similarity computation, they typically fail to incorporate context information for accurate depth estimation in ambiguous areas. To tackle this issue, some approaches~\cite{mayer2016large, kendall2017end, chang2018pyramid} take advantage of contextual information from features of various scales in an end-to-end manner. Specifically, Mayer \etal~\cite{mayer2016large} first propose an end-to-end network that deploys a 1D correlation layer called DispNetC. Nonetheless, most stereo networks still inherently struggle to handle textureless, reflective and strongly occluded areas in the stereo images.
\vspace{5pt}

\begin{figure*}[!t]
\begin{center}
\includegraphics[width=0.88\linewidth]{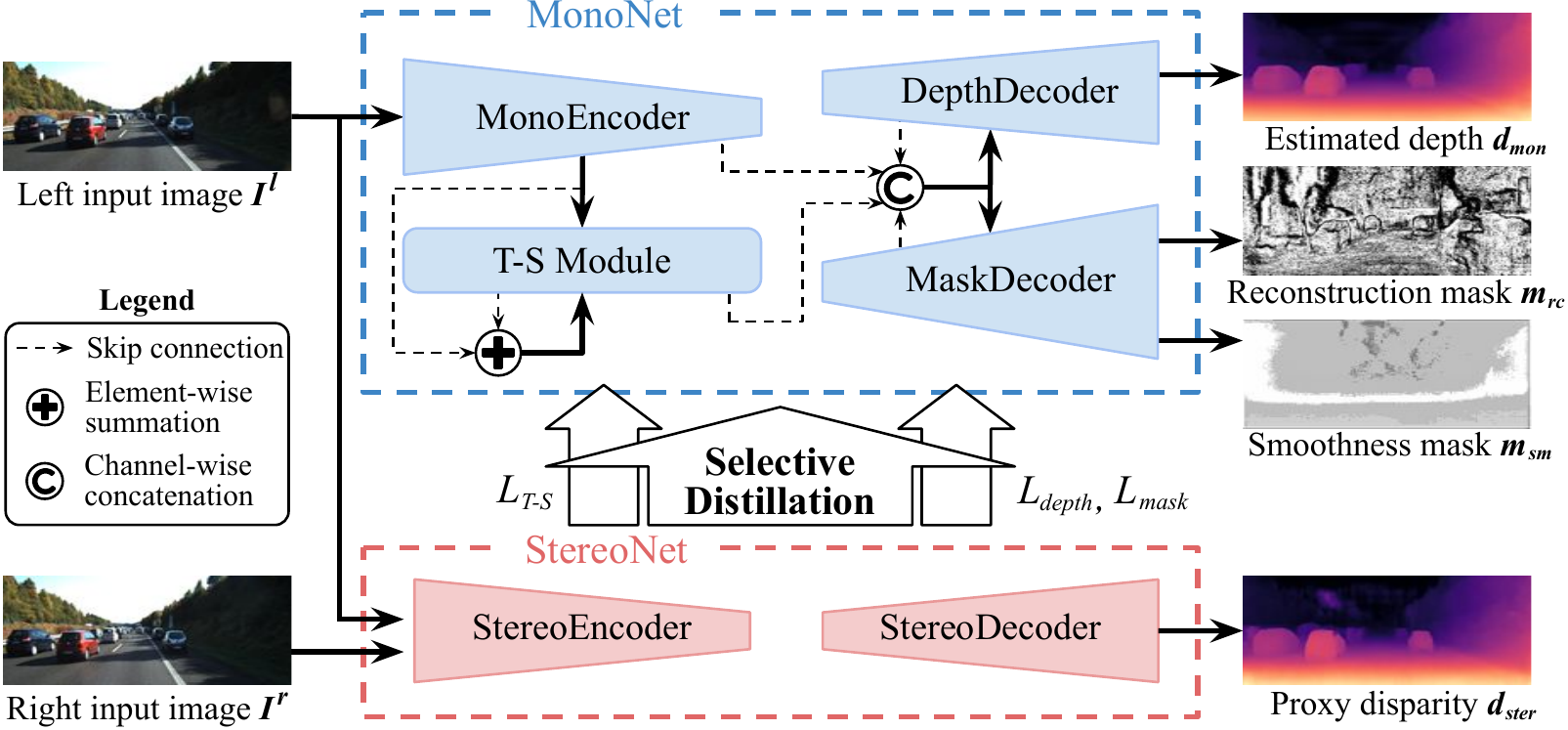}
\end{center}
\vspace{-10pt} \captionsetup{font=small}
    \caption{Overview of our training framework. Given pairs of stereo images, our monocular network (\textit{MonoNet}) aims to recover depth maps $d_{mon}$ and learn two binary masks $m_{rc}$ and $m_{sm}$ which \textbf{selectively distill} disparity maps $d_{ster}$ predicted by a pre-trained stereo matching network (\textit{StereoNet}). The \textit{T-S module} further improves the MonoNet by distilling geometric features from the StereoEncoder. Please note that only a single image is required during evaluation. Best viewed in color.}
\label{fig:overall_framework}
\end{figure*}

\noindent\textbf{\noindent{Monocular depth estimation}} \hspace{1mm}
In contrast to the stereo-based methods, monocular depth estimation infers depth information from only a single image, yet an inherently ill-posed problem since pixels in the image can have multiple possible depths. Nevertheless, with the availability of ground-truth depth maps, CNN-based supervised learning methods~\cite{alhashim2018high, lee2019monocular, gan2018monocular, fang2020towards, lee2019big, fu2018deep} have recently been sought. 
Although they have achieved remarkable results in accuracy, they often stuck in undesirable local optima. Moreover, obtaining labeled data or annotating ground-truths is costly in itself. Alternatively, numerous self-supervised methods have been proposed to overcome the dependence on the ground-truths. Garg \etal~\cite{garg2016unsupervised} first formulate the self-supervised monocular depth estimation through photometric coherence between stereo images. Godard \etal~\cite{godard2017unsupervised} further improve~\cite{garg2016unsupervised} by imposing a left-right consistency to the estimated disparity maps. Poggi \etal~\cite{poggi2018learning} more recently recast the training strategy of~\cite{godard2017unsupervised} to assume that three images aligned horizontally are available at training phase, overcoming the limitations caused by the binocular setup. Godard \etal~\cite{godard2019digging} 
have extended their previous work~\cite{godard2017unsupervised} by their novel reprojection loss, auto-masking strategy and upsampling its depth estimations to the input resolution. Besides, a variety of monocular depth estimation methods have been explored with the aid of semantic segmentation~\cite{chen2019towards}, self-attention module~\cite{chen2019attention, johnston2020self}, camera intrinsic parameters~\cite{gordon2019depth} and generative adversarial networks~\cite{aleotti2018generative, cs2018monocular, atapour2018real}, to name a few.
\vspace{5pt}

\noindent\textbf{\noindent{Proxy-supervised depth estimation}} \hspace{1mm}
No matter how the monocular depth estimation methods have been advanced, stereo-based methods are still generally known to provide better accuracy and have better generalization ability to other domains during evaluation than most monocular depth estimation networks. As a way of leveraging these advantages, a few recent works have suggested utilizing disparity maps from stereo vision as proxy ground-truths instead of expensive ground-truths to learn monocular depth estimation. Tosi \etal~\cite{tosi2019learning} suggest utilizing the SGM to generate disparity maps and distilling it as proxy labels for the proxy-supervised loss as well as the self-supervised loss for monocular depth estimation. Watson \etal~\cite{watson2019self} propose to generate a number of disparity maps by SGM to use them as depth hints for the network to escape from local minima. Since both works have utilized the traditional stereo methods, it is less time-consuming and provides more disparity maps than learning-based stereo methods. Rather than utilizing conventional methods(\textit{i.e}, non-learning methods), Guo \etal~\cite{guo2018learning} propose to distill stereo matching networks pre-trained on synthetic datasets, aiming to improve the generalization ability of monocular depth estimation network. They propose to use the disparity maps predicted by the pre-trained stereo network as proxy ground truths to supervise their monocular depth estimation network. However, they do not take account of the fact that stereo-based methods often fail to provide proper supervision to monocular depth estimation methods in some areas. We extend their idea in a way that is particularly designed for our purpose, namely, selective distillation of stereo knowledge.

\section{Proposed methods}
\label{sec:proposed_methods}
The goal of our methods is to estimate dense depth maps from a single image without any ground-truth depth maps such that are obtained by depth sensors (\eg LiDARs), by distilling the stereo network only where the monocular depth estimation network needs to learn from it. This section first provides an overall pipeline of our framework and then describes the details of the proposed methods to further enhance the accuracy of depth estimation by selectively distilling the stereo knowledge.
\subsection{Overall pipeline}
\label{sec:overall_pipeline}
As illustrated in Fig.~\ref{fig:overall_framework}, our network consists of two sub-networks: \textit{MonoNet} (top) and \textit{StereoNet} (bottom). We make full use of large quantities of easy-to-acquire synthetic datasets by distilling disparity maps predicted by the StereoNet pre-trained on synthetic datasets since it is less desirable to train the MonoNet directly on synthetic datasets. We specifically adopt the pre-trained stereo networks from Guo \etal~\cite{guo2018learning} as our StereoNet, since they have already verified the feasibility of using their stereo networks as a proxy supervisory network, though any other pre-trained stereo networks can be adopted. Since the size of features extracted from the StereoNet of each scale differs from that of the MonoNet's features, they are resized to fit the size of the features from MonoNet at each scale for distillation. Please note that the StereoNet is never used during evaluation and none of its parameter is updated while training the MonoNet, since it is included in our framework only for the purpose of proxy distillation.

Meanwhile, our MonoNet is designed with a standard encoder-decoder architecture and extra T-S module as Fig.~\ref{fig:overall_framework} depicts. Specifically, we adopt VGG-16~\cite{simonyan2014very} pretrained on ImageNet~\cite{russakovsky2015imagenet} as our MonoEncoder and it extracts the initial feature volume of the same spatial resolution as the input image ($256\times512$), with the channel depth of $32$, and then successively reduces the spatial resolution by half and doubles the channel depth. After then, as shown in Fig.~\ref{fig:overall_framework}, features extracted from the MonoEncoder pass through the extra convolutional layers (\textit{T-S module}), turning into student features detailed in Sec.~\ref{sec:teacher_student_module}. Finally, the DepthDecoder learns to estimate multi-scale monocular depth maps while the MaskDecoder learns two different binary masks for selective distillation, discussed in detail in Sec.~\ref{sec:selective_distillation}. 

\subsection{Selective distillation}
\label{sec:selective_distillation}
To train our MonoNet, the StereoNet, pre-trained with a large amount of synthetic data,  is utilized to provide proxy ground truths. A few related works have utilized the disparity maps predicted by stereo-based methods as proxy ground truths to directly supervise the monocular depth maps by using L1 loss or reverse Huber (berHu) loss. However, we note that the MonoNet should learn from the StereoNet only when it is required to, rather than carelessly learn from every pixel. For example, it is generally recognized that stereo-based methods show more weakness in occluded areas(\textit{i.e.} textureless, repeated patterns) or close distance~\cite{martins2018fusion, gil2019monster} than monocular methods. Therefore, the MonoNet needs to discriminate and select only those areas where the proxy supervision is valid.

The MaskDecoder is designed to learn two binary masks that give selection criteria for the monocular depth estimation network to selectively distill the proxy disparity maps. Since there is no ground truth for these two binary masks, we use the commonly-used loss functions for self-supervised monocular depth estimation, image reconstruction loss and edge-aware smoothness loss. For the image reconstruction loss, the reference image $I^{l}$ is reconstructed in a way that the other image $I^{r}$ is warped to the reference viewpoint based on the predicted disparity $d$, and then the photometric discrepancy between $I^{l}$ and ${I}^{r}(d)$ is measured as:
\vspace{-12pt}
\begin{equation}
\label{eqn:image_reconstruction_loss}
\begin{split}
L_{rc}(I^{l}, I^{r}(d)) &= \frac{1}{N}\sum_{i,j}\alpha\frac{1-ZNCC(I_{ij}^{l},I_{ij}^{r}(d_{ij}))}{2}\\&+(1-\alpha)||I_{ij}^{l}-I_{ij}^{r}(d_{ij})||.\\[-3pt]
\end{split}
\end{equation}
Although the Structural Similarity(SSIM) has been generally used as a patch similarity measurement in an image reconstruction loss, we adopt the Zero mean Normalized Cross Correlation (ZNCC) with $3\times3$ patch and $\alpha$ set to 0.85, instead of the SSIM, following~\cite{chen2018self, kim2020loop}. Meanwhile, the predicted disparity is constrained to be locally smooth by the edge-aware smoothness loss as: 
\begin{equation}
\label{eqn:smoothness_loss}
\vspace{-20pt}
L_{sm}(I^{l}, d) = \frac{1}{N}\sum_{i,j}\left|\partial_{x} d_{ij}\right| e^{-\left|\partial_{x} I_{ij}^{l}\right|}+\left|\partial_{y} d_{ij}\right| e^{-\left|\partial_{y} I_{ij}^{l}\right|}.
\end{equation} \vspace{-12pt}

\begin{figure}[!t]
\begin{center}
\includegraphics[width=1.0\linewidth]{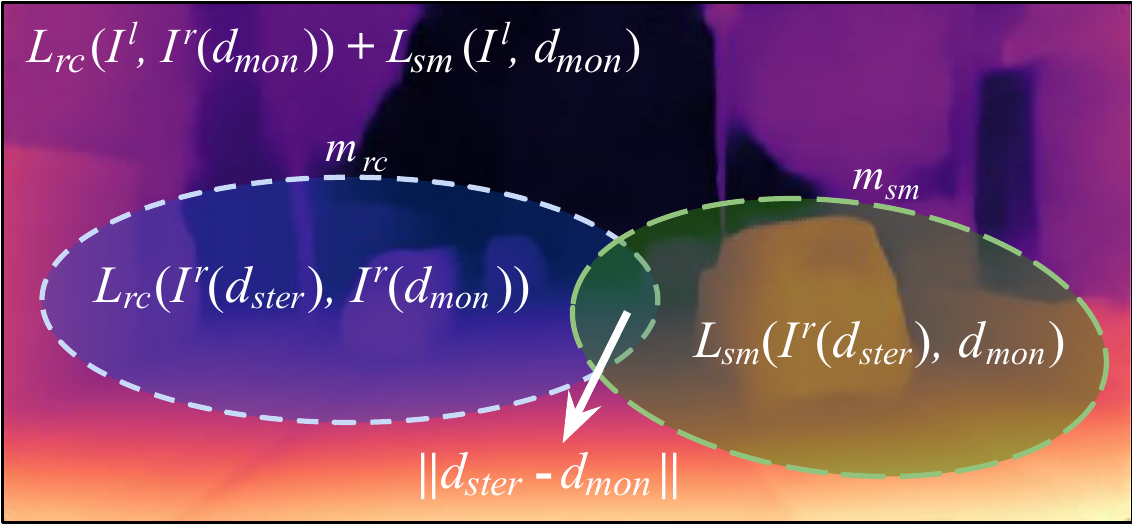}
\end{center}
\vspace{-14pt} \captionsetup{font=small}
   \caption{Illustration of $L_{depth}$ with the two binary masks on the estimated depth $d_{mon}$. Each mask is valid in a different area and the intersected area is equal to the element-wise multiplication of $m_{rc}$ and $m_{sm}$. Best viewed in color.}
\label{fig:mask_diagram}
\end{figure}

Each of the two binary masks $m_{rc}$ and $m_{sm}$ learns to form new virtual disparity maps $d_{rc}$ and $d_{sm}$, respectively, by selecting between the proxy disparity map $d_{ster}$ and currently-estimated monocular depth map $d_{mon}$ for each pixel, as follows:
\begin{gather}
\label{eqn:virtual_disparity_map_formation}
\vspace{-20pt}
d_{rc} = \sum_{i,j}m_{rc(ij)}d_{ster(ij)}+(1-m_{rc(ij)})d_{mon(ij)}\\
d_{sm} = \sum_{i,j}m_{sm(ij)}d_{ster(ij)}+(1-m_{sm(ij)})d_{mon(ij)}.
\vspace{-20pt}
\end{gather}
The virtual disparity maps $d_{rc}$ and $d_{sm}$ are then used to calculate  $L_{rc}$ and $L_{sm}$, respectively. For instance, $m_{rc}$ aims to learn to optimally select between $d_{ster}$ and $d_{mon}$ to form $d_{rc}$ that minimizes $L_{rc}$. Hence, the total loss for MaskDecoder is as follows:
\begin{equation}
\label{eqn:mask_loss}
\vspace{-20pt}
L_{mask} = L_{rc}(I^{l}, I^{r}(d_{rc}))+L_{sm}(I^{l}, d_{sm}).
\vspace{-20pt}
\end{equation}

\begin{figure}[!ht]
\begin{center}
\includegraphics[width=1.0\linewidth]{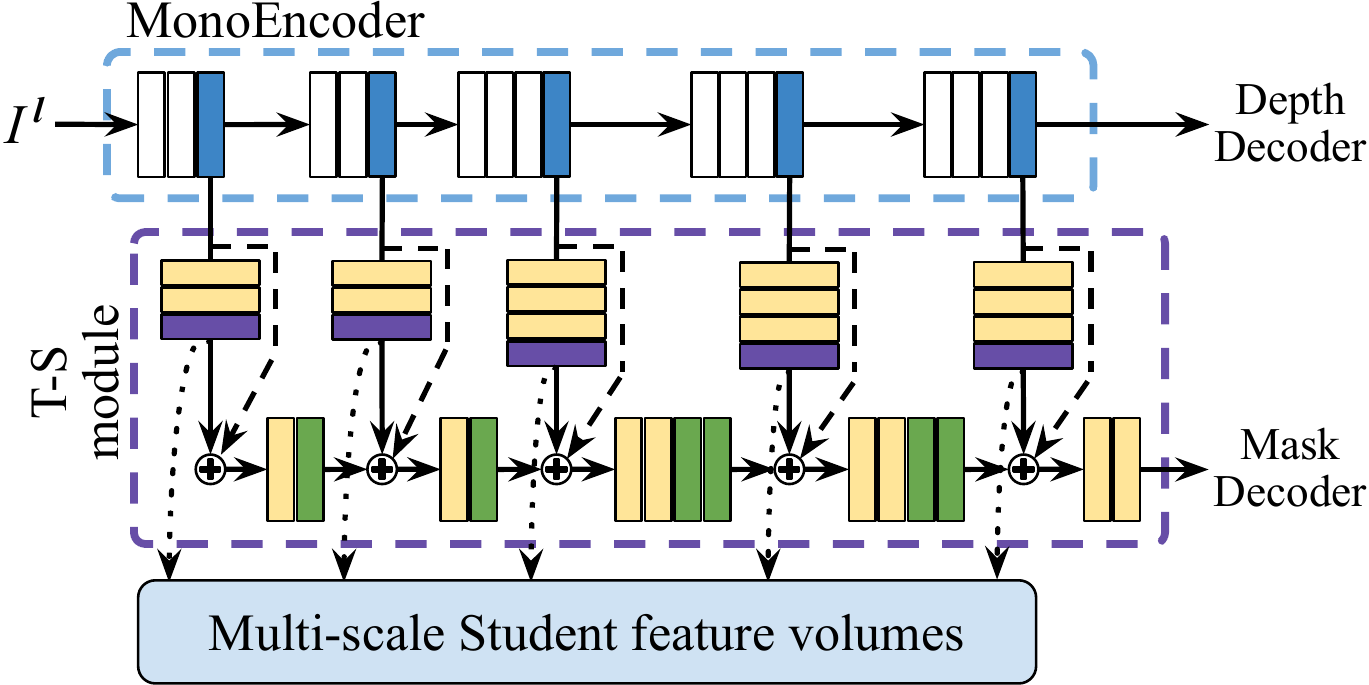}
\end{center}
\vspace{-12pt} \captionsetup{font=small}
   \caption{Structure of our MonoEncoder and T-S module.
   The MonoEncoder extracts features at multi-scale and then feeds them to the T-S module to form student features. The student features at each scale are supervised by the features extracted by the StereoEncoder. 
   The legend in Fig.~\ref{fig:overall_framework} also applies to this figure.}
\vspace{-6pt}
\label{fig:mononet_architecture}
\end{figure}

The two binary masks are then utilized to train the DepthDecoder. As well as the two common loss functions of Eq.~\ref{eqn:image_reconstruction_loss} and~\ref{eqn:smoothness_loss}, the DepthDecoder is guided by the proxy disparity map $d_{ster}$, being conscious of the areas where it needs to improve upon (\textit{i.e.} $m_{rc}$ and $m_{sm}$), as follows:
\begin{equation}
\label{eqn:mask_loss}
\begin{split}
\vspace{-20pt}
L_{depth} &= L_{rc}(I^{l}, I^{r}(d_{mon}))+m_{rc}L_{rc}(I^{r}(d_{ster}), I^{r}(d_{mon}))\\
&+L_{sm}(I^{l}, d_{mon})+m_{sm}L_{sm}(I^{r}(d_{ster}), d_{mon})\\
&+(m_{rc}\times m_{sm})||d_{ster}-d_{mon}||. \\[-18pt]
\end{split}
\end{equation}
$(m_{rc} \times m_{sm})$ is interpreted as a region where the both criteria agree that $d_{ster}$ is superior to $d_{mon}$. In those reliable areas, we further enforce the proxy supervision by using a L1 loss as illustrated in Fig.~\ref{fig:mask_diagram}.

\subsection{T-S module}
\label{sec:teacher_student_module}
\noindent\textbf{Structure}\hspace{1mm}
It is reasonable to learn monocular depth estimation utilizing the geometric cues of the features encoded by the StereoNet. However, since our framework requires a single image during testing, it is unreliable to feed the features extracted from the stereo image pair directly to the MonoNet. Alternatively, we design a T-S module so that our MonoNet indirectly exploits geometric features from the learned student features. First, as Fig.~\ref{fig:mononet_architecture} shows, each multi-scale feature volume extracted by the MonoEncoder undergoes additional convolutional layers and is made into a student feature volume of the same size as each feature volume from the Mono/StereoEncoder. This T-S module aggregates the features from the MonoEncoder and the newly generated student features through element-wise summation at each scale, followed by extra convolutions. The white and blue blocks indicate 3$\times$3 convolution layer and 2D max pooling, respectively. The yellow, purple and green blocks in the T-S module indicate 3$\times$3 convolution layer, 1$\times$1 convolution with batch normalization, and 3$\times$3 convolution of stride 2, respectively. ReLU is used as an activation function following each convolution in the MonoEncoder while Leaky ReLU is utilized in the T-S Module. The two different feature volumes extracted from the MonoEncoder and the T-S module at each scale are then concatenated with other feature volumes from the Depth/MaskDecoder in a channel-wise manner as Fig.~\ref{fig:overall_framework} shows.

\vspace{3pt}

\noindent\textbf{T-S loss functions}\hspace{1mm}
To distill the teacher features into the student features, we propose to utilize three kinds of distillation loss functions. We first define the Feature Distillation loss based on the simple L2 distance norm:
\begin{equation}
\vspace{-20pt}
L_{FD} = \sum_{i=1}^{4}\sum_{c=1}^{C_{i}}\sum_{h=1}^{H_{i}}\sum_{w=1}^{W_{i}}\frac{{0.5}^{i-1}}{c\times h\times w}||F_{T}^{i} - F_{S}^{i}||_2  
\vspace{-20pt}
\end{equation}
where $F^{i}\in R^{C_{i}\times H_{i}\times W_{i}}$ denotes the feature map extracted by an encoder, $C_{i}$ and ($H_{i}$, $W_{i}$) the number of output channels and spatial dimensions of $F^{i}$ at $i$-th scale, respectively. The subscripts $T$ and $S$ indicate whether each feature map is extracted by a Teacher (\textit{StereoEncoder}) or a Student (\textit{MonoEncdoer}).

Additionally, to take into account the fact 
that each channel has different weights of geometrical cues, the Channel Distillation loss, $L_{CD}$, is utilized for the student features to learn the each channel weight of the teacher features. The weight of $c$-th channel $wt^{c}$ and the Channel Distillation loss $L_{CD}$ are defined as follows: 
\vspace{-4pt}
\begin{gather}
\vspace{-15pt}
wt^{c} = \sum_{i=1}^{4}\sum_{h=1}^{H_{i}}\sum_{w=1}^{W_{i}}\frac{F^{i}}{h\times w} \\[4pt]
L_{CD} = \sum_{i=1}^{4}\sum_{c=1}^{C_i}\frac{||wt_{T}^{c}-wt_{S}^{c}||_2}{c}.
\vspace{-15pt}
\end{gather}\\[-4pt]
$L_{CD}$ is designed based on the simple L2 distance between the \emph{channel-wise mean} of various scale feature volumes extracted from the teacher encoder and the student encoder, while $L_{FD}$ can be interpreted as the mean L2 distance between the feature volumes of teacher and student.

Inspired by \cite{tung2019similarity, peng2019correlation}, we lastly suggest to use the Similarity Distillation loss, $L_{SD}$, which is based on the cosine similarity function. First we flatten $F^{i}$ into $Q^{i} \in R^{C_{i}\times H_{i}W_{i}}$, and then derive the tensor $Z^{i}$ = $Q^{i}\cdot {Q^{i}}^\top$, which are normalized to unit length row-by-row, as $\tilde{Z^{i}}$ = $Z^{i} / ||Z^{i}||_2$. The Similarity Distillation loss $L_{SD}$ is then defined as
\vspace{-6pt}
\begin{equation}
L_{SD} = \sum_{i=1}^{4}\frac{||\tilde{Z_{T}^{i}}-\tilde{Z_{S}^{i}}||_2}{C_{i}^{2}}.
\end{equation} \\[-10pt]
$L_{SD}$ aims to penalize cosine dissimilarity between two feature maps, constraining the student feature maps to semantically resemble the teacher feature maps. Hence, the total loss function for the T-S feature distillation is defined as follows:
\vspace{-5pt}
\begin{equation}
L_{T-S} = L_{FD}+\lambda_{CD}L_{CD}+\lambda_{SD}L_{SD}.
\vspace{-25pt}
\end{equation}

\begin{figure*}[t]
\begin{center}
\includegraphics[width=1\linewidth]{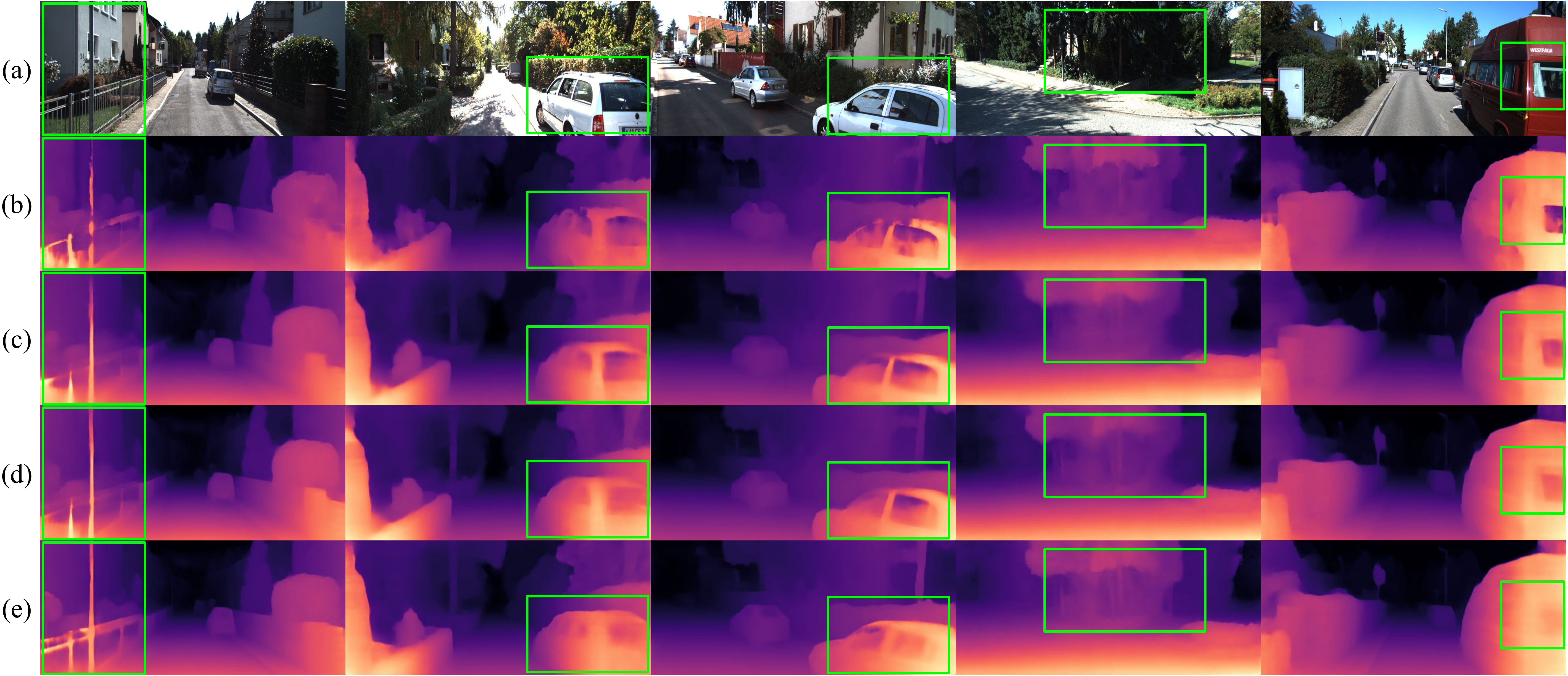}
\end{center}
\vspace{-13pt} \captionsetup{font=small}
   \caption{
   (a) Given training images, the pre-trained StereoNet (S,K$\rightarrow$K) outputs (b) proxy disparity maps. (c) Guo~\etal~\cite{guo2018learning} learn some areas where they should not have learned (the green boxes) from the proxy, while (d) ours with \textit{selective distillation} even improves upon it. (e) Additionally leveraging the T-S module brings more sharp and accurate depth estimation. Best viewed in color.}
\label{fig:distillation_comparison}
\end{figure*}

\noindent$\lambda_{CD}$ and $\lambda_{SD}$ balance the loss terms, empirically set to $1$ and $10^{5}$, respectively. 
The effects of each T-S distillation loss function on depth estimation accuracy are discussed in Sec.~\ref{subsec:ablation_study}.

To sum up, the entire network is trained to optimize the total loss function, $L_{total}$, consisting of three losses $L_{depth}$, $L_{mask}$ and $L_{T-S}$ as follows:
\begin{equation}
\label{eqn:total_loss}
\vspace{-10pt}
L_{total} = L_{depth}+\lambda_{mask}L_{mask}+\lambda_{T-S}L_{T-S}.
\vspace{-10pt}
\end{equation}
We empirically set the balancing-parameters $\lambda_{mask}$ and $\lambda_{T-S}$ to be $1$ and $10^{-4}$ respectively. Please note $\lambda_{T-S}$ is set to be $0$ for our model without the T-S module.

\section{Experiments}
\label{sec:experiments}
In this section, we first give the details of experimental setup. We then provide the ablation studies on the proposed methods that affect the monocular depth estimation accuracy. Lastly, we evaluate our framework on the Eigen test split of the KITTI dataset~\cite{geiger2013vision} and compare it to other state-of-the-art self-, proxy-, and semi-supervised methods.

\begin{table*}[!t]
	\begin{center} \captionsetup{font=small}
	    \caption{Quantitative results of the ablation study on the selective distillation(rows 1\&2 and 7\&8) and the T-S module combined with various T-S loss functions(rows 3-6). S and K in the dataset column denote the Scene Flow dataset~\cite{mayer2016large} and the KITTI dataset~\cite{geiger2013vision} respectively. All the results are evaluated on the Eigen test split of the KITTI dataset. The results of the baseline model, with none of our contributions, are from Guo~\etal~\cite{guo2018learning}. \textbf{The best performance} is \textbf{boldfaced} and \uline{the second best} is \uline{underlined}. The same notation and typography are applied to the following tables.}
	    \vspace{-5pt}
		\scalebox{0.86}
		{
			\begin{tabular}{cc|c|c|c|c|cccc|ccc}
				\hline
				\multicolumn{1}{c}{\multirow{2}{*}{Datasets}} &
				\multirow{2}{*}{\begin{tabular}[c]{@{}c@{}}Distillation\\ Method\end{tabular}} & \multirow{2}{*}{\begin{tabular}[c]{@{}c@{}}T-S \\ Module\end{tabular}} & \multirow{2}{*}{\begin{tabular}[c]{@{}c@{}}$L_{FD}$\end{tabular}} & \multirow{2}{*}{\begin{tabular}[c]{@{}c@{}}$L_{CD}$\end{tabular}} &	\multirow{2}{*}{\begin{tabular}[c]{@{}c@{}}$L_{SD}$\end{tabular}} &
				\multicolumn{4}{c|}{Lower the better} & \multicolumn{3}{c}{Higher the better} \\
				\multicolumn{1}{c}{} &  &  &  &  &  & Abs. Rel. & Sq. Rel. & RMSE & RMSE log & $\delta < 1.25$ & $\delta < 1.25^2$ & $\delta < 1.25^3$ \\ \hline
				S,K$\rightarrow$ K & Direct~\cite{guo2018learning} &  &  &  &  & 0.101 & 0.690 & 4.254 & 0.173 & 0.884 & \uline{0.966} & \textbf{0.986} \\
				S,K$\rightarrow$ K & Selective  &  &  &  &  & 0.099 & 0.668 & 4.212 & 0.172 & 0.888 & \uline{0.966} & \textbf{0.986} \\
				S,K$\rightarrow$ K & Selective & \checkmark & \checkmark &  &  & 0.100 & 0.688 & 4.225 & 0.173 & 0.888 & 0.965 & \textbf{0.986} \\
				S,K$\rightarrow$ K & Selective & \checkmark & \checkmark & \checkmark &  & 0.098 & 0.691 & 4.197 & \uline{0.171} & 0.891 & \uline{0.966} & \textbf{0.986} \\
				S,K$\rightarrow$ K & Selective & \checkmark & \checkmark &  & \checkmark & \uline{0.097} & \uline{0.666} & \uline{4.169} & \textbf{0.169} & \uline{0.893} & \textbf{0.967} & \textbf{0.986} \\
				S,K$\rightarrow$ K & Selective & \checkmark & \checkmark & \checkmark & \checkmark & \textbf{0.093} & \textbf{0.650} & \textbf{4.147} & \textbf{0.169} & \textbf{0.897} & \textbf{0.967} & \textbf{0.986} \\ \hline
				S$\rightarrow$ K & Direct~\cite{guo2018learning} &  &  &  &  & 0.109 & 0.822 & 4.656 & \uline{0.192} & 0.868 & 0.958 & 0.981 \\
				S$\rightarrow$ K & Selective  &  &  &  &  & \uline{0.100} & \textbf{0.719} & \textbf{4.387} & \textbf{0.182} & \uline{0.883} & \uline{0.962} & \textbf{0.984} \\
				S$\rightarrow$ K & Selective & \checkmark & \checkmark & \checkmark & \checkmark & \textbf{0.098} & \uline{0.729} & \uline{4.411} & \textbf{0.182} & \textbf{0.889} & \textbf{0.963} & \uline{0.983} \\ \hline
			\end{tabular}
			}
		\label{tab:ablation_study}
	\end{center}
	\vspace{-8pt}
\end{table*}

\subsection{Experimental setup}
\label{subsec:experimental_setup}
\vspace{-3pt}
\noindent\textbf{Datasets}\hspace{1mm}
The two main stereo datasets of the KITTI~\cite{geiger2013vision} and the CityScapes~\cite{cordts2016cityscapes} datasets are used to train and evaluate our framework. First, the KITTI dataset contains 61 outdoor scenes of rectified stereo pairs, captured from a moving car equipped with a LiDAR. For fair comparisons, we exploit the Eigen split~\cite{eigen2014depth} of the KITTI dataset containing 22,600 and 697 image pairs for training and testing respectively, with the standard cap of 80m~\cite{godard2017unsupervised}. On the other hand, the CityScapes dataset consisting of 22,973 urban stereo pairs is used only to pre-train our MonoNet which is finetuned on the KITTI dataset afterwards. We cut off the bottom part of the CityScapes image since most of the images include the car hood. Please note that we do not elaborate the details of the Scene Flow dataset~\cite{mayer2016large} in this section, although it is used as the only synthetic dataset for training the proxy stereo networks. That is because we have particularly adopted the two pre-trained stereo networks from~\cite{guo2018learning} as our StereoNet throughout the whole experiments; a StereoNet which is trained on the Scene Flow dataset alone (S$\rightarrow$K) and another which is subsequently finetuned on 100 scenes from the KITTI 2015 dataset with ground truth depth maps (S,K$\rightarrow$K). S and K denote the Scene Flow and the KITTI datasets, respectively.

\vspace{3pt}

\noindent\textbf{Evaluation metrics}\hspace{1mm}
Our framework is evaluated by the standard evaluation metrics for monocular depth estimation as following: absolute relative error (Abs. Rel.), squared relative error (Sq. Rel.), root mean squared error (RMSE), root mean squared logarithmic error (RMSE log) and accuracy $\delta$ with thresholds of [$1.25$, $1.25^2$, $1.25^3$]. $\delta$ represents the percentage of predicted depths $d_{i}$ which meet the condition that its maximum between ratio and inverse ratio with respect to the ground-truth depths $g_{i}$ is lower than a threshold, mathematically expressed as $\delta$=$max$($d_{i} \over g_{i}$, $g_{i} \over d_{i}$).

\vspace{3pt}

\noindent\textbf{Implementation details}\hspace{1mm}
During training, the Adam optimizer~\cite{kingma2014adam} is used with the parameters $\beta_{1}$ = $0.9$, $\beta_{2}$ = $0.999$ and $\epsilon$ = $10^{-8}$. The input images and the proxy disparity maps are fed to our MonoNet after resized to the size of $256\times512$ and are augmented following the same way as~\cite{guo2018learning}. The MonoNet is trained for 50 epochs with the initial learning rate of $10^{-4}$ halved at epoch 20, 35, 45. The batch size is set as 8 and 12 for the model with and without the T-S module respectively, in consideration of the number of parameters. In terms of network implementation, the MonoEncoder and T-S module extracts features at 5 scales while the Depth and MaskDecoder predicts depth and mask at 4 scales. Please see supplementary material for a detailed demonstration of the connections among the encoder, T-S module and decoders. Note the CityScapes training set is utilized to pre-train the MonoNet in the same manner as the KITTI training set is used.

\subsection{Ablation study}
\label{subsec:ablation_study}
\noindent\textbf{Selective distillation}\hspace{1mm}
As mentioned in Sec.~\ref{sec:selective_distillation}, each of monocular and stereo-based methods shows superior estimation accuracy to each other in different regions. Along with this, Fig.~\ref{fig:distillation_comparison} depicts the StereoNet shows poor depth estimation on some areas where it is hard to find matching correspondences, such as textureless or reflective areas (\eg shadowed area or car windows), and areas where disparity discontinuity is repeated (\eg thin fences). Since~\cite{guo2018learning} is trained to directly distill the disparity maps without taking account of the aforementioned consideration, their monocular depth estimation network is supervised even in those areas where the proxy supervision is unreliable. As opposed to~\cite{guo2018learning}, the proposed selective distillation method has shown to successfully prevent training with the unreliable supervision from proxy disparity maps, while even improving the accuracy in areas where the monocular depth estimation network is superior to the stereo network.

The quantitative improvement on performance is demonstrated in Table~\ref{tab:ablation_study}. When we compare rows 1\&2 and 7\&8, our selective distillation method has achieved more accurate performance than the direct distillation method~\cite{guo2018learning} in every evaluation metric on the Eigen test split of the KITTI dataset. In particular, comparison between rows 7 and 8 verifies that our MonoNet achieves more accurate performance on the real dataset (KITTI) even though the proxy stereo network is trained only on the synthetic datasets (S$\rightarrow$K). Note that it has improved more significantly than the case the proxy stereo network is finetuned on the real dataset. From this, it can be said that our selective distillation method becomes more crucial when the proxy stereo network is less reliable to distill.

\begin{table*}[!ht]
	\begin{center} \captionsetup{font=small}
	    \caption{Quantitative comparison with the state-of-the-arts on the Eigen test split of the KITTI dataset. ``pp" indicates the post-processing is applied to the results. Please note that ours even surpasses the semi-supervised methods with the LiDAR ground truth depth maps. The same notation and typography are applied to the following table.}
	    \vspace{-5pt}
		\scalebox{0.88}{
			\begin{tabular}{lc|c|cccc|ccc}
				\hline
				\multicolumn{1}{c}{\multirow{2}{*}{Method}} &
				\multirow{2}{*}{\begin{tabular}[c]{@{}c@{}}Dataset\end{tabular}} &	\multirow{2}{*}{\begin{tabular}[c]{@{}c@{}}Supervision\end{tabular}} &
				\multicolumn{4}{c|}{Lower the better} & \multicolumn{3}{c}{Higher the better} \\
				\multicolumn{1}{c}{} &  &  & Abs. Rel. & Sq. Rel. & RMSE & RMSE log & $\delta < 1.25$ & $\delta < 1.25^2$ & $\delta < 1.25^3$ \\ \hline
				Godard~\etal~\cite{godard2017unsupervised} & K & Self & 0.148 & 1.344 & 5.927 & 0.247 & 0.803 & 0.922 & 0.964 \\
				Wong~\etal~\cite{wong2019bilateral} & K & Self  & 0.133 & 1.126 & 5.515 & 0.231 & 0.826 & 0.934 & 0.969 \\
				Poggi~\etal~\cite{poggi2018learning}, pp & K & Self & 0.126 & 0.961 & 5.205 & 0.220 & 0.835 & 0.941 & 0.974 \\
				Chen~\etal~\cite{chen2019towards} & K & Self & 0.118 & 0.905 & 5.096 & 0.211 & 0.839 & 0.945 & 0.977 \\
				Guizilini~\etal~\cite{guizilini20203d} & K & Self & 0.111 & 0.785 & 4.601 & 0.189 & 0.878 & 0.960 & 0.982 \\	Johnston~\etal~\cite{johnston2020self} & K & Self & 0.106 & 0.861 & 4.699 & 0.185 & \textbf{0.889} & \uline{0.962} & 0.982 \\
				Tosi~\etal~\cite{tosi2019learning} & K & Proxy(SGM) & 0.111 & 0.867 & 4.714 & 0.199 & 0.864 & 0.954 & 0.979 \\
				Watson~\etal~\cite{watson2019self} & K & Proxy(SGM) & 0.102 & 0.762 & 4.602 & 0.189 & 0.880 & 0.960 & 0.981 \\
				Guo~\etal~\cite{guo2018learning} & S$\rightarrow$ K & Proxy(StereoNet) & 0.109 & 0.822 & 4.656 & 0.192 & 0.868 & 0.958 & 0.981 \\
				Ours(w/o T-S module) & S$\rightarrow$ K & Proxy(StereoNet) & \uline{0.100} & \textbf{0.719} & \textbf{4.387} & \textbf{0.182} & \uline{0.883} & \uline{0.962} & \textbf{0.984} \\
				Ours(Full model) & S$\rightarrow$ K & Proxy(StereoNet) & \textbf{0.098} & \uline{0.729} & \uline{4.411} & \textbf{0.182} & \textbf{0.889} & \textbf{0.963} & \uline{0.983} \\ \hline
				Guo~\etal~\cite{guo2018learning} & S,K$\rightarrow$ K & Proxy(StereoNet) & 0.101 & 0.690 & 4.254 & 0.173 & 0.884 & \uline{0.966} & \textbf{0.986} \\
				Ours(w/o T-S module) & S,K$\rightarrow$ K & Proxy(StereoNet) & \uline{0.099} & \uline{0.668} & \uline{4.212} & \uline{0.172} & \uline{0.888} & \uline{0.966} & \textbf{0.986} \\
				Ours(Full model) & S,K$\rightarrow$ K & Proxy(StereoNet) & \textbf{0.093} & \textbf{0.650} & \textbf{4.147} & \textbf{0.169} & \textbf{0.897} & \textbf{0.967} & \textbf{0.986} \\ \hline
				Kuznietsov~\etal~\cite{kuznietsov2017semi} & K & Semi(LiDAR GT) & 0.113 & 0.741 & 4.621 & 0.189 & 0.862 & 0.960 & 0.986 \\
				Yang~\etal~\cite{yang2018deep} & K & Semi(LiDAR GT) & 0.097 & 0.734 & 4.442 & 0.187 & 0.888 & 0.958 & 0.980 \\ \hline
			\end{tabular}}
		\label{tab:comparison_sota_kitti}
	\end{center}
	\vspace{-8pt}
\end{table*}

\begin{table*}[!hbt]
	\begin{center} \captionsetup{font=small}
	    \caption{Quantitative comparison with the state-of-the-arts pre-trained on the CityScapes dataset (C) and then finetuned on the KITTI dataset (K). All models are evaluated on the Eigen test split of the KITTI dataset. ``*" in the Supervision column indicates the models are supervised by monocular sequences and the corresponding results are from the paper of~\cite{tosi2019learning}.}
	    \vspace{-5pt}
		\scalebox{0.87}{
			\begin{tabular}{lc|c|cccc|ccc}
				\hline
				\multicolumn{1}{c}{\multirow{2}{*}{Method}} &
				\multirow{2}{*}{\begin{tabular}[c]{@{}c@{}}Dataset\end{tabular}} &	\multirow{2}{*}{\begin{tabular}[c]{@{}c@{}}Supervision\end{tabular}} &
				\multicolumn{4}{c|}{Lower the better} & \multicolumn{3}{c}{Higher the better} \\
				\multicolumn{1}{c}{} &  &  & Abs. Rel. & Sq. Rel. & RMSE & RMSE log & $\delta < 1.25$ & $\delta < 1.25^2$ & $\delta < 1.25^3$ \\ \hline
				Mahjourian~\etal~\cite{mahjourian2018unsupervised} & C,K & Self* & 0.159 & 1.231 & 5.912 & 0.243 & 0.784 & 0.923 & 0.970 \\
				
				Wang~\etal~\cite{wang2018learning} & C,K & Self* & 0.148 & 1.187 & 5.496 & 0.226 & 0.812 & 0.938 & 0.975 \\
				
				Zou~\etal~\cite{zou2018df} & C,K & Self* & 0.146 & 1.182 & 5.215 & 0.213 & 0.818 & 0.943 & 0.978 \\
				
				Wong~\etal~\cite{wong2019bilateral} & C,K & Self & 0.118 & 0.996 & 5.134 & 0.215 & 0.849 & 0.945 & 0.975 \\
				Godard~\etal~\cite{godard2017unsupervised}, pp & C,K & Self  & 0.114 & 0.898 & 4.935 & 0.206 & 0.861 & 0.949 & 0.976 \\
				Poggi~\etal~\cite{poggi2018learning}, pp & C,K & Self & 0.111 & 0.849 & 4.822 & 0.202 & 0.865 & 0.952 & 0.978 \\
				Pilzer~\etal~\cite{pilzer2019refine} & C,K & Self & 0.098 & 0.831 & 4.656 & 0.202 & 0.882 & 0.948 & 0.973 \\
				Tosi~\etal~\cite{tosi2019learning} & C,K & Proxy(SGM) & 0.096 & 0.673 & 4.351 & 0.184 & 0.890 & 0.961 & 0.981 \\
				Guo~\etal~\cite{guo2018learning} & S,K$\rightarrow$ C,K & Proxy(StereoNet) & \uline{0.096} & \textbf{0.641} & 4.095 & 0.168 & 0.892 & \uline{0.967} & \uline{0.986} \\
				Ours(w/o T-S module) & S,K$\rightarrow$ C,K & Proxy(StereoNet) & \textbf{0.095} & 0.661 & \uline{4.088} & \uline{0.166} & \uline{0.897} & \textbf{0.969} & \textbf{0.987} \\
				Ours(Full model) & S,K$\rightarrow$ C,K & Proxy(StereoNet) & \textbf{0.095} & \uline{0.654} & \textbf{4.073} & \textbf{0.165} & \textbf{0.900} & \textbf{0.969} & \textbf{0.987} \\ \hline
			\end{tabular}}
		\label{tab:comparison_sota_cityscapes}
	\end{center}
	\vspace{-15pt}
\end{table*}

\vspace{5pt}
\noindent\textbf{T-S loss functions}\hspace{1mm}
To verify the validity of the proposed T-S module, we have conducted an ablation study by applying different combinations of the three different T-S loss functions, $L_{FD}$, $L_{CD}$ and $L_{SD}$. Since the Feature Distillation loss $L_{FD}$ has been set as the base loss function of the total T-S loss function, we first compare our model \textit{without} the T-S module to that \textit{with} the T-S module combined with $L_{FD}$ alone. The results in Table~\ref{tab:ablation_study} (rows 2 and 3) demonstrate that applying $L_{FD}$ alone fails to properly transfer the stereo feature representations from the Teacher to the Student. We hypothesize this is because the volume of the feature is too large for the student to learn adequately. On the other hand, when $L_{CD}$ is additionally applied (row 4), the performance on every metric except Sq. Rel. has been slightly improved. It can be said $L_{CD}$ successfully enables to supervise the large number of channels in the student feature volume. Moreover, the addition of $L_{SD}$ (row 5) brings larger improvement than when $L_{CD}$ is additionally applied to $L_{FD}$. This is because $L_{SD}$ successfully constrains the student features to learn semantic and geometrical knowledge from the teacher features. When combined with all the proposed T-S loss functions, our full model (rows 6 and 9) achieves significant improvement over the baseline model with the directive distillation method. Meanwhile, when we compare rows 2\&6 and 8\&9, the T-S module is much more effective when the Teacher network (StereoNet) is more reliable to distill, as apposed to the selective distillation.


\subsection{Comparison with the state-of-the-arts}

In this section, we compare our framework with the state-of-the-art self-, proxy-, and semi-supervised monocular depth estimation methods. In Table~\ref{tab:comparison_sota_kitti}, ours and other different models are evaluated on the Eigen test split of the KITTI dataset. From the table, it can be noticed ours outperforms all the others significantly. In particular, even our model leveraging the StereoNet trained only on the synthetic dataset (S$\rightarrow$K) (rows 10 and 11) outperforms the semi-supervised methods supervised by the ground truth depth maps~\cite{kuznietsov2017semi, yang2018deep} in most metrics. From these results, it can be said selectively distilling the proxy disparity maps allows to achieve even better performance than using the expensive ground truths. Besides, ours also significantly surpasses all the other state-of-the-art methods supervised in different settings.

In Table~\ref{tab:comparison_sota_cityscapes}, we compare the performances of our model with other state-of-the-art monocular depth estimation models pre-trained on the CityScapes dataset before being finetuned on the KITTI dataset. We first train our MonoNet on the CityScapes dataset by selectively distilling a StereoNet that is trained only on the Scene Flow dataset(S$\rightarrow$C), and then finetune the pre-trained MonoNet on the Eigen train split of the KITTI dataset by distilling another StereoNet that is finetuned on the KITTI dataset (S,K$\rightarrow$K). We can see the results have improved over using the KITTI dataset alone. However, our full model has been slightly improved and even performances on some metrics declined. Nonetheless, our full model still outperforms all the other state-of-the-art models except on the Sq. Rel. metric.


\section{Conclusion}
\vspace{-5pt}
\label{sec:conclusion}
In this paper, we pointed out that distillation of proxy disparity maps from stereo-based methods for monocular depth estimation should be selectively exploited. 
With this consideration, we proposed a novel framework for learning monocular depth estimation via selective distillation of the stereo knowledge to tackle the problems of the previous proxy-supervised monocular depth estimation methods.
Qualitative results proved our selective distillation during training enables to avoid unreliable supervision of the proxy disparity map. In addition, extensive experiments demonstrated the T-S module combined with the various distillation loss functions further improves our network to achieve state-of-the-art performance of proxy-supervised monocular depth estimation methods, even outperforming some of the semi-supervised methods on the KITTI dataset.

{\small
\bibliographystyle{ieee_fullname}
\bibliography{egpaper}
}

\end{document}